\documentclass{article} %
\usepackage[accepted]{icml2020}[]

\usepackage{times}
\usepackage{hyperref}
\usepackage{url}
\usepackage{graphicx}

\usepackage{lipsum}
\usepackage{textcomp}

\usepackage{listings}
\usepackage{textgreek} %
\usepackage{amsmath} %

\usepackage[capitalise]{cleveref}
\usepackage{stfloats} 
\usepackage{float}
\usepackage{placeins} %

\usepackage{tablefootnote} %
\usepackage[roman]{parnotes} %
\usepackage{tabulary}
\usepackage{booktabs}
\usepackage{tabularx}

\usepackage{changepage}

\definecolor{mygray}{gray}{0.5}
\definecolor{cblue}{RGB}{8, 85, 153}
\definecolor{darkblue}{RGB}{1, 43, 112}

\definecolor{cgreen}{RGB}{8, 153, 83}
\definecolor{green}{RGB}{8, 200, 50}
\definecolor{red}{RGB}{255, 0, 0}

\renewcommand{\algorithmiccomment}[1]{\hfill \textcolor{mygray}{\# #1}}
\renewcommand\algorithmicthen{:}
\renewcommand\algorithmicdo{:}

\newcommand{\pos}[1]{{\textcolor{green}{\textbf{#1}}}}
\newcommand{\nega}[1]{{\textcolor{red}{\textbf{#1}}}}

\newcommand{\fref}[1]{Fig~\ref{#1}}
\newcommand{\eref}[1]{Eq~\ref{#1}}
\newcommand{\sref}[1]{Sec~\ref{#1}}
\newcommand{\tref}[1]{Table~\ref{#1}}

\usepackage{amsfonts}

\usepackage[definitionLists,hashEnumerators,smartEllipses,hybrid]{markdown} %

\icmltitlerunning{Interpretations are Useful: Penalizing Explanations to Align Neural Networks with Prior Knowledge}

\begin{document} 

\twocolumn[
\icmltitle{Interpretations are Useful: \\Penalizing Explanations to Align Neural Networks with Prior Knowledge}
\begin{icmlauthorlist}
\icmlauthor{Laura Rieger}{dtu}
\icmlauthor{Chandan Singh}{ucbe}
\icmlauthor{W. James Murdoch}{ucb}
\icmlauthor{Bin Yu}{ucbe,ucb}
\end{icmlauthorlist}
\icmlcorrespondingauthor{Laura Rieger}{lauri@dtu.dk}

\icmlaffiliation{dtu}{DTU Compute, Technical University Denmark, 2800  Kgs. Lyngby, Denmark}
\icmlaffiliation{ucb}{Department of Statistics, UC Berkeley, Berkeley, California, USA}
\icmlaffiliation{ucbe}{EECS Department, UC Berkeley, Berkeley, California, USA}
\vskip 0.3in
]
\printAffiliationsAndNotice{}

\begin{abstract}

    For an explanation of a deep learning model to be effective, it must both provide insight into a model and suggest a corresponding action in order to achieve an objective. Too often, the litany of proposed explainable deep learning methods stop at the first step, providing practitioners with insight into a model, but no way to act on it. In this paper we propose contextual decomposition explanation penalization (CDEP), a method that enables practitioners to leverage explanations to improve the performance of a deep learning model. In particular, CDEP enables inserting domain knowledge into a model to ignore spurious correlations, correct errors, and generalize to different types of dataset shifts. We demonstrate the ability of CDEP to increase performance on an array of toy and real datasets.
\end{abstract}

\section{Introduction}

In recent years, deep neural networks (DNNs) have demonstrated strong predictive performance across a wide variety of settings. However, in order to predict accurately, they sometimes latch onto spurious correlations caused by dataset bias or overfitting \citep{winkler2019association}. Moreover, DNNs are also known to exploit bias regarding gender, race, and other sensitive attributes present in training datasets \citep{GargEmbeddings,obermeyer2019dissecting,dressel2018accuracy}. Recent work in explaining DNN predictions \citep{murdoch2019interpretable, doshi2017towards} has demonstrated an ability to reveal the relationships learned by a model. Here, we extend this line of work to not only uncover learned relationships, but penalize them to improve a model.

\begin{figure}[th!]
	\vskip 0.2in
	\begin{center}
		\centerline{\includegraphics[width=\columnwidth]{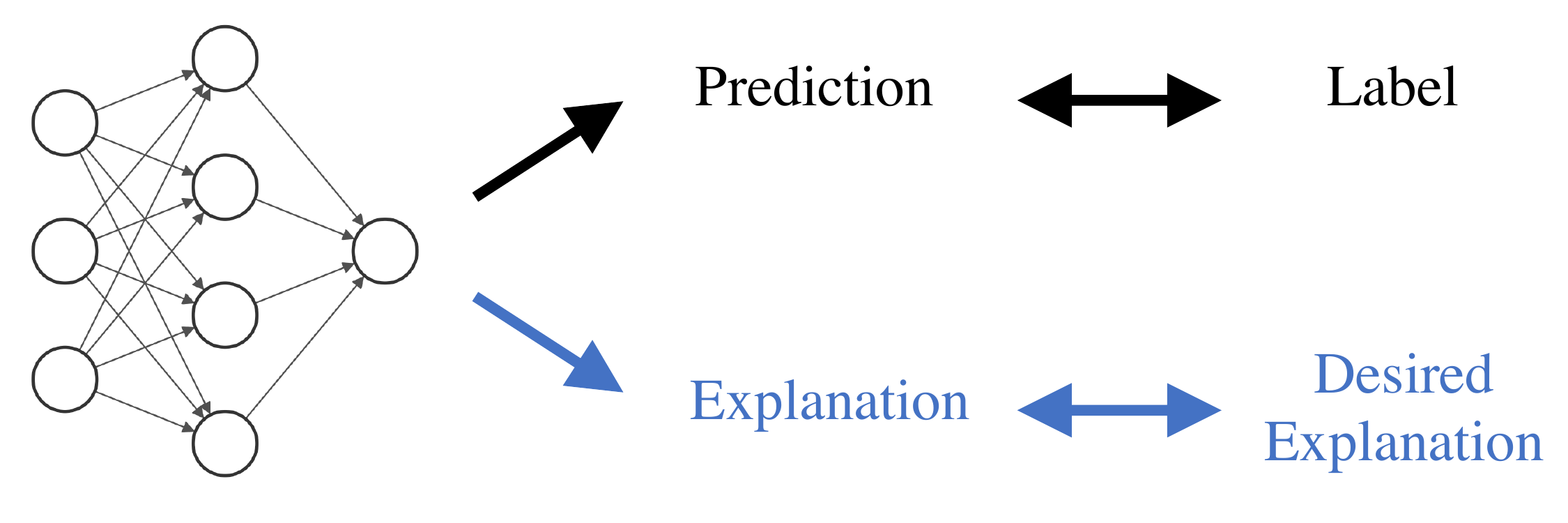}}
		\caption{CDEP allows practitioners to penalize both a model's prediction and the corresponding explanation.}
		\label{fig:intro}
	\end{center}
	\vskip -0.2in
\end{figure}

We introduce \underline{c}ontextual \underline{d}ecomposition \underline{e}xplanation \underline{p}enalization (CDEP), a method which leverages a particular existing explanation technique for neural networks to enable the insertion of domain knowledge into a model. Given prior knowledge in the form of importance scores, CDEP works by allowing the user to directly penalize importances of certain features or feature interactions. This forces the neural network to not only produce the correct prediction, but also the correct explanation for that prediction.\footnote{ Code, notebooks, scripts, documentation, and models for reproducing experiments here and using CDEP on new models available at \url{https://github.com/laura-rieger/deep-explanation-penalization}.}

While we focus on the use of contextual decomposition, which allows the penalization of both feature importances and interactions \citep{murdoch2018beyond, singh2018hierarchical}, CDEP can be readily adapted for existing interpretation techniques, as long as they are differentiable. Moreover, CDEP is a general technique, which can be applied to arbitrary neural network architectures, and is often orders of magnitude faster and more memory efficient than recent gradient-based methods, allowing its use on meaningful datasets.

We demonstrate the effectiveness of CDEP via experiments across a wide array of tasks. In the prediction of skin cancer from images, CDEP improves the prediction of a classifier by teaching it to ignore spurious confounders present in the training data. 

In a variant of the MNIST digit-classification task where the digit's color is used as a misleading signal, CDEP regularizes a network to focus on a digit's shape rather than its color. 
Finally, simple examples show how CDEP can help mitigate fairness issues, both in text classification and risk prediction.

\section{Background}
\label{sec:background}

\paragraph{Explanation methods} Many methods have been developed to help explain the learned relationships contained in a DNN. For local or prediction-level explanation, most prior work has focused on assigning importance to individual features, such as pixels in an image or words in a document. There are several methods that give feature-level importance for different architectures. They can be categorized as 
 gradient-based \citep{springenberg2014striving, sundararajan2016gradients, selvaraju2016grad, baehrens2010explain, rieger2019aggregating}, 
 decomposition-based \citep{murdoch2017automatic, shrikumar2016not, bach2015pixel}
 and others \citep{dabkowski2017real, fong2017interpretable, ribeiro2016should,zintgraf2017visualizing}, 
 with many similarities among the methods \citep{ancona2018towards, lundberg2017unified}. 
However, many of these methods have  been poorly evaluated so far \citep{adebayo2018sanity, nie2018theoretical}, casting doubt on their usefulness in practice. Another line of work, which we build upon, has focused on uncovering interactions between features \citep{murdoch2018beyond}, and using those interactions to create a hierarchy of features displaying the model's prediction process \citep{singh2019hierarchical,ch2020transformation}. 

\paragraph{Uses of explanation methods} 

While much work has been put into developing methods for explaining DNNs, relatively little work has explored the potential to use these explanations to help build a better model.
Some recent work proposes forcing models to attend to regions of the input which are known to be important \citep{burns2018women, mitsuhara2019embedding}, although it is important to note that attention is often not the same as explanation \citep{jain2019attention}. 

An alternative line of work proposes penalizing the gradients of a neural network to match human-provided binary annotations and shows the possibility to improve performance \citep{ross2017right,bao2018deriving,du2019learning} and adversarial robustness \citep{ross2018improving}. Two recent papers extend these ideas by penalizing  gradient-based attributions for natural language models \citep{liu2019incorporating} and to produce smooth attributions \citep{erion2019learning}. \citet{du2019learning} applies a similar idea to improve image segmentation by incorporating attention maps into the training process. 

Predating deep learning, \citeauthor{zaidan2007using} (\citeyear{zaidan2007using}) consider the use of ``annotator rationales'' in sentiment analysis to train support vector machines. This work on annotator rationales was recently extended to show improved explanations (not accuracy) in particular types of CNNs \citep{strout2019human}.

\paragraph{Other ways to constrain DNNs} 
While we focus on the use of explanations to constrain the relationships learned by neural networks, other approaches for constraining neural networks have also been proposed. A computationally intensive alternative is to augment the dataset in order to prevent the model from learning undesirable relationships, through domain knowledge \citep{bolukbasi2016man}, projecting out superficial statistics \citep{wang2019learning} or dramatically altering training images \citep{geirhos2018imagenet}. However, these processes are often not feasible, either due to their computational cost or the difficulty of constructing such an augmented data set. Adversarial training has also been explored \citep{zhang2019interpreting}. These techniques are generally limited, as they are often tied to particular datasets, and do not provide a clear link between learning about a model's learned relationships through explanations, and subsequently correcting them. 
\section{Methods}
\label{sec:methods}

In the following, we will first establish the general form of the augmented loss function. We then describe Contextual Decomposition (CD), the explanation method proposed by \citep{murdoch2018beyond}.  Based on this, we introduce CDEP and point out its desirable computational properties for regularization. In \cref{subsec:right_knowledge} we describe how prior knowledge can be encoded into explanations and give examples of typical use cases. 
While we focus on CD scores, which allow the penalization of interactions between features in addition to features themselves, our approach readily generalizes to other interpretation techniques, as long as they are differentiable.

\subsection{Augmenting the loss function}

Given a particular classification task, we want to teach a model to not only produce the correct prediction but also to arrive at the prediction for the correct reasons.  That is, we want the model to be right for the right reasons, where the right reasons are provided by the user and are dataset-dependent. Assuming a truthful explanation method, this implies that the explanation provided by the DNN for a particular decision should be aligned with a pre-supplied explanation encoding our knowledge of the underlying reasons. 

To accomplish this, we augment the traditional objective function used to train a neural network, as displayed in \eref{eq:general_method} with an additional component. In addition to the standard prediction loss $\mathcal{L}$, which teaches the model to produce the correct predictions by penalizing wrong predictions, we  add an explanation error $\mathcal{L}_{\text{expl}}$, which teaches the model to produce the correct explanations for its predictions by penalizing wrong explanations. 

In place of the prediction and labels $f_\theta(X), y$, used in the prediction error $\mathcal{L}$, the explanation error $\mathcal{L}_{\text{expl}}$ uses the explanations produced by an interpretation method $\text{expl}_\theta(X)$, along with targets provided by the user $\text{expl}_X$. As is common with penalization, the two losses are weighted by a hyperparameter $\lambda \in \mathbb{R}$:

\begin{equation}
\begin{split}
\label{eq:general_method}
\hat{\theta} = \underset{\theta}{\text{argmin}} \: &\overbrace{\mathcal{L}\left(f_\theta(X), y\right)}^{\text{Prediction error}} \\
+ \lambda  &\underbrace{\mathcal{L}_{\text{expl}}\left(\text{expl}_\theta(X), \text{expl}_X\right)}_{\text{Explanation error}}
\end{split}
\end{equation}

The precise meaning of $\text{expl}_X$ depend on the context. For example, in the skin cancer image classification task described in \cref{sec:results}, many of the benign skin images contain band-aids, while none of the malignant images do. To force the model to ignore the band-aids in making their prediction, in each image $\text{expl}_\theta(X)$ denotes the importance score of the band-aid and $\text{expl}_X$ would be zero. These and more examples are further explored in \cref{sec:results}.

\subsection{Contextual decomposition (CD)} 

In this work, we use the CD score as the explanation function. In contrast to other interpretation methods, which focus on feature importances, CD also captures interactions between features, making it particularly suited to regularize the importance of complex features.

CD was originally designed for LSTMs \citep{murdoch2018beyond} and subsequently extended to convolutional neural networks and arbitrary DNNs \citep{singh2018hierarchical}.  For a given DNN $f(x)$, one can represent its output as a SoftMax operation applied to logits $g(x)$.  These logits, in turn, are the composition of $L$ layers $g_i$, such as convolutional operations or ReLU non-linearities.
\begin{align}
f(x) &= \text{SoftMax}(g(x)) \\ &= \text{SoftMax}(g_L(g_{L-1}(...(g_2(g_1(x))))))
\end{align}

Given a group of features $\{x_j\}_{j \in S}$, the CD algorithm, $g^{CD}(x)$, decomposes the logits $g(x)$ into a sum of two terms, $\beta(x)$ and $\gamma(x)$. $\beta(x)$ is the importance score of the feature group $\{x_j\}_{j \in S}$, and $\gamma(x)$ captures contributions to $g(x)$ not included in $\beta(x)$. The decomposition is computed by iteratively applying decompositions $g^{CD}_i(x)$ for each of the layers $g_i(x)$.
\begin{align}
 g^{CD}(x) &= g^{CD}_L(g^{CD}_{L-1}(...(g^{CD}_2(g^{CD}_1(x)))))) \\
 &= (\beta(x), \gamma(x)) \\
 & = g(x)
\end{align}

\subsection{CDEP objective function} 

We now substitute the above CD scores into the generic equation in \eref{eq:general_method} to arrive at CDEP as it is used in this paper. While we use CD for the explanation method $\text{expl}_\theta(X)$, other explanation methods could be readily substituted at this stage. In order to convert CD scores to probabilities, we apply a SoftMax operation to $g^{CD}(x)$, allowing for easier comparison with the user-provided labels $\text{expl}_X$. We collect from the user, for each input $x_i$, a collection of feature groups $x_{i, S}$, $x_i \in \mathbb{R}^d, S \subseteq \{1,...,d\}$, along with explanation target values $\text{expl}_{x_{i, S}}$, and use the $\|\cdot \|_1$ loss for $\mathcal{L}_{\text{expl}}$.

This yields a vector $\beta(x_j)$ for any subset of features in an input $x_j$ which we would like to penalize. We can then collect ground-truth label explanations for this subset of features, $\text{expl}_{x_j}$ and use it to regularize the explanation. Using this we arrive at the equation for the weight parameters with CDEP loss:

\begin{equation}
\begin{split}
\hat{\theta} = \underset{\theta}{\text{argmin}} \: &\overbrace{ \underset{i}{\sum} \underset{c}{\sum} -y_{i, c} \log f_{\theta}(x_i)_{c}}^{\text{Prediction error}} 
\\ +  \lambda & \underbrace{\sum_i\sum_S  ||\beta(x_{i, S}) - \text{expl}_{x_{i, S}}||_1}_{\text{Explanation error}}
\end{split}
\label{eq:specific_methods}
\end{equation}

In the above, $i$ indexes each individual example in the dataset, $S$ indexes a subset of the features for which we penalize their explanations, and $c$ sums over each class. 

Updating the model parameters in accordance with this formulation ensures that the model not only predicts the right output but also does so for the right (aligned with prior knowledge) reasons. It is important to note that the evaluation of what the right reasons are depends entirely on the practitioner deploying the model. As with the class labels, using wrong or biased explanations will yield a wrong and biased model.

\subsection{Encoding domain knowledge as explanations} 
\label{subsec:right_knowledge}
The choice of ground-truth explanations $\text{expl}_X$ is dependent on the application and the existing domain knowledge. 
CDEP allows for penalizing arbitrary interactions between features, allowing the incorporation of a very broad set of domain knowledge.

In the simplest setting, practitioners may precisely provide groundtruth human explanations for each data point. This may be useful in a medical image classifications setting, where data is limited and practitioners can endow the model with knowledge of how a diagnosis should be made. However, collecting such groundtruth explanations can be very expensive.

To avoid assigning human labels, one may utilize programmatic rules to identify and assign groundtruth importance to regions, which are then used to help the model identify important/unimportant regions. For example, \sref{subsec:isic_results} uses rules to identify spurious patches in images which should have zero importance and \sref{subsec:sst} uses rules to identify and assign zero importance to words involving gender.

In a more general case, one may specify importances of different feature interactions. For example in \sref{subsec:mnist} we specify that the importance of pixels in isolation should be zero, so only interactions between pixels can be used to make predictions. This prevents a model from latching onto local cues such as color and texture when making its prediction.

\subsection{Computational considerations} 
Previous work has proposed ideas similar to \eref{eq:general_method}, where the choice of explanation method is based on gradients \citep{ross2017right, erion2019learning}. However, using such methods leads to three main complications which are solved by our approach.

The first complication is the optimization process. When optimizing over gradient-based attributions via gradient descent, the optimizer requires the gradient of the gradient,  requiring that all network components be twice differentiable. This process is computationally expensive and optimizing it exactly involves optimizing over a differential equation, often making it intractable. In contrast, CD attributions are calculated along the forward pass of the network, and as a result, can be optimized plainly with back-propagation using the standard single forward-pass and backward-pass per batch.

A second advantage from the use of CD in \eref{eq:specific_methods} is the ability to quickly finetune a pre-trained network. In many applications, particularly in transfer learning, it is common to finetune only the last few layers of a pre-trained neural network. Using CD, one can freeze early layers of the network and quickly finetune final layers, as the calculation of gradients of the frozen layers is not necessary.

Third, CDEP incurs much lower memory usage than competing gradient-based methods. With  gradient-based methods the training requires the storage of activations and gradients for all layers of the network as well as the gradient with respect to the input (which can be omitted in normal training). Even for the simplest gradient-based methods, this more than doubles the required memory for a given batch and network size, sometimes becoming prohibitively large. In contrast, penalizing CD requires only a small constant amount of memory more than standard training.

\section{Results}
\label{sec:results}

The results here demonstrate the efficacy of CDEP on a variety of datasets using diverse explanation types. \sref{subsec:isic_results} shows results on ignoring spurious patches in the ISIC skin cancer dataset \citep{codella2019skin}, \sref{subsec:mnist} details experiments on converting a DNN's preference for color to a preference for shape on a variant of the MNIST dataset \citep{lecun1998mnist}, \sref{ssec:compas} showcases the use of CDEP to train a  neural network that aligns better with a pre-defined fairness measure, and \sref{subsec:sst} shows experiments on text data from the Stanford Sentiment Treebank (SST) \citep{socher2013recursive}.\footnote{All models were trained in PyTorch \cite{paszke2017automatic}.}

\subsection{Ignoring spurious signals in skin cancer diagnosis}
\label{subsec:isic_results}
\begin{figure*}[!hbtp]
	\vskip 0.2in
	\begin{center}
		\includegraphics[width=\textwidth]{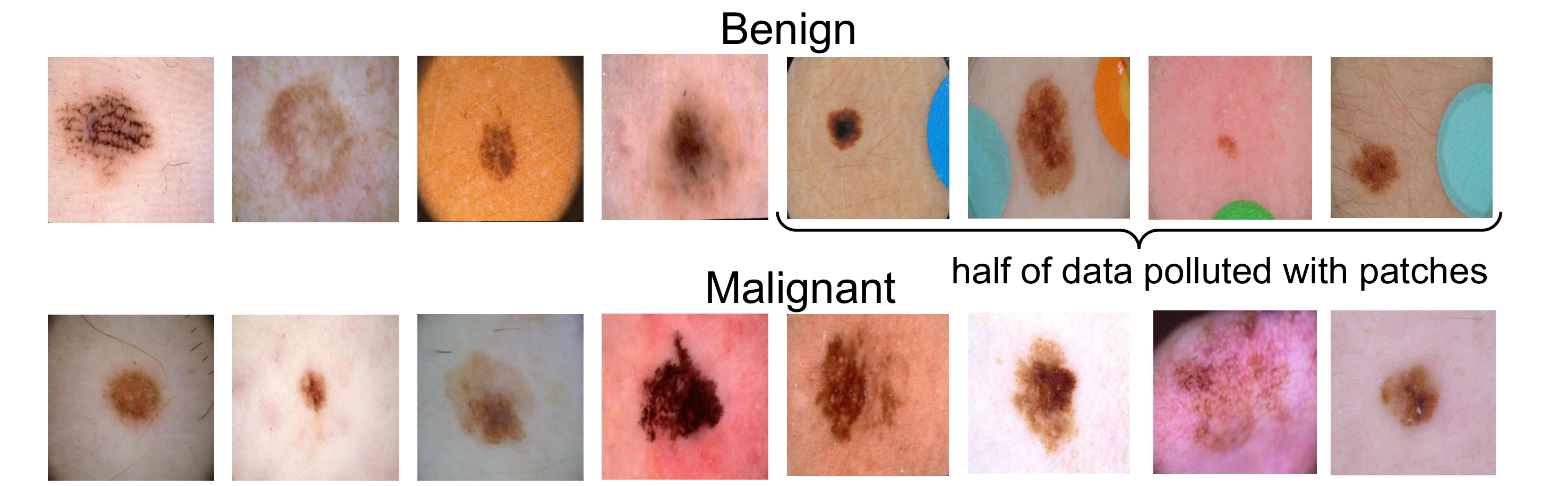}
	\end{center}
	\caption{Example images from the ISIC dataset. Half of the benign lesion images include a patch in the image. Training on this data results in the neural network overly relying on the patches to classify images. We aim to avoid this with our method.
	}
	\vskip -0.2in
	\label{fig:example_ISIC}
\end{figure*}

\begin{figure}[tbp!]
	\vskip 0.2in
	\begin{center}
		\centerline{\includegraphics[width=1\columnwidth]{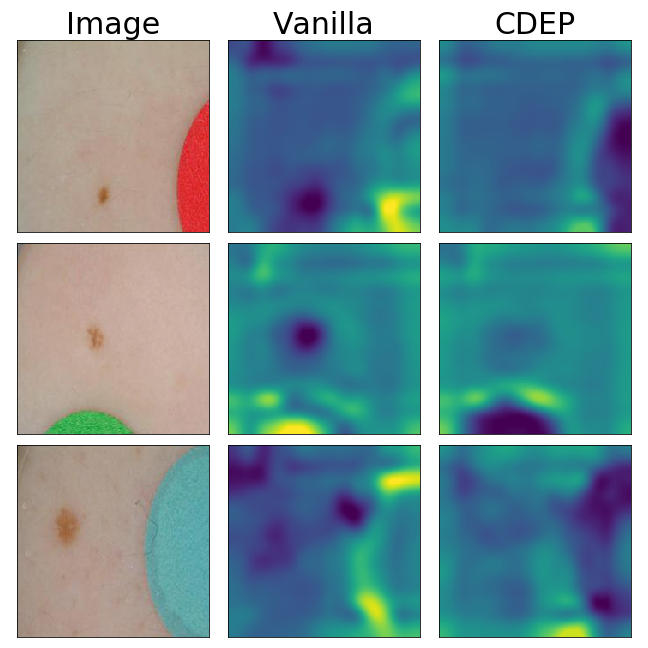}}
		\caption{Visualizing heatmaps for correctly predicted exampes from the ISIC skin cancer test set. Lighter regions in the heatmap are attributed more importance. The DNN trained with CDEP correctly captures that the patch is not relevant for classification. }
		\label{fig:gradcam}
	\end{center}
	\vskip -0.2in
\end{figure}

In recent years, deep learning has achieved impressive results in diagnosing skin cancer, with predictive accuracy sometimes comparable to human doctors \citep{esteva2017dermatologist}. 
However, the datasets used to train these models often include spurious features which make it possible to attain high test accuracy without learning the underlying phenomena \citep{winkler2019association}. In particular, a popular dataset from ISIC (International Skin Imaging Collaboration) has colorful patches present in approximately 50\% of the non-cancerous images but not in the cancerous images as can be seen in \cref{fig:example_ISIC} \citep{codella2019skin,tschandl2018ham10000}. An unpenalized DNN learns to look for these patches as an indicator for predicting that an image is benign as can be seen in \cref{fig:gradcam}. We use CDEP to remedy this problem by penalizing the DNN placing importance on the patches during training.

The task in this section is to classify whether an image of a skin lesion contains (1) benign lesions or (2) malignant lesions. In a real-life task, this would for example be done to determine whether a biopsy should be taken.
 The ISIC dataset consists of 21,654 images with a certain diagnosis (19,372 benign, 2,282 malignant), each diagnosed by histopathology or a consensus of experts. We excluded 2247 images since they had an unknown or not certain diagnosis.

To obtain the binary maps of the patches for the skin cancer task, we first segment the images using SLIC, a common image-segmentation algorithm \citep{achanta2012slic}. Since the patches are a different color from the rest of the image, they are usually their own segment. 
Subsequently we take the mean RGB and HSV values for all segments and filter for segments in which the mean was substantially different from the typical caucasian skin tone. Since different images were different from the typical skin color in different attributes, we filtered for those images recursively. As an example, in the image shown in the appendix in Fig. S3, the patch has a much higher saturation than the rest of the image.

After the spurious patches were identified, we penalized them with CDEP  to have zero importance. For classification, we use a VGG16 architecture \citep{simonyan2014very} pre-trained on the ImageNet Classification task\citep{imagenet_cvpr09}\footnote{Pre-trained model retrieved from \href{https://pytorch.org/docs/stable/torchvision/models.html}{\textcolor{cblue}{torchvision}}.} and freeze the weights of early layers so that only the fully connected layers are trained. To account for the class imbalance present in the dataset, we weigh the classes to be equal in the loss function. 

\cref{tab:ISIC_results} shows results comparing the performance of a model trained with and without CDEP. We report results on two variants of the test set. The first, which we refer to as ``no patches'' only contains images of the test set that do not include patches. The second also includes images with those patches. Training with CDEP improves the AUC and F1-score for both test sets.
\begin{table*}[hbtp!]

	\caption{Results from training a DNN on ISIC to recognize skin cancer (averaged over three runs). Results shown for the entire test set and for only the images the test set that do not include patches (``no patches''). The network trained with CDEP generalizes better, getting higher AUC and F1 on both. Std below $0.006$ for all AUC and below $0.012$ for all F1.}
	\label{tab:ISIC_results}
	\begin{center}
	\begin{small}
\begin{sc}
\begin{tabular}{lrrrrr}
\toprule
{} & AUC (no patches) & F1 (no patches) & AUC (all) & F1 (all) \\
& & & & & \\
\midrule
Vanilla (Excluding training data with patches) & 0.88 & 0.59 & 0.93 & 0.58  \\
Vanilla & 0.87 & 0.56 & 0.93 & 0.56 \\
RRR & {0.75 }& {0.46} & {0.86} & {0.44} \\
CDEP & \textbf{0.89 }& \textbf{0.61} & \textbf{ 0.94} & \textbf{0.60} \\
\bottomrule
\end{tabular}
\end{sc}
\end{small}
	\end{center}
\end{table*}

In the first row of \cref{tab:ISIC_results}, the model is trained using only the data without the spurious patches, and the second row shows the model trained on the full dataset. The network trained using CDEP achieves the best F1 score, surpassing both unpenalized versions.

Interestingly, the model trained with CDEP also improves when we consider the entire (biased) dataset, indicating that the model does in fact generalize better to all examples. 
We also compared our method against the method introduced in 2017 by \citeauthor{ross2017right} (RRR). For this, we restricted the batch size to 16 (and consequently use a learning rate of $10^{-5}$) due to memory constraints.\footnote{A higher learning rate yields NaN loss and a higher batch size requires too much GPU RAM, necessitating these settings. Due to this a wider sweep of hyperparameters was not possible.}

Using RRR did not improve on the base AUC, implying that penalizing gradients is not helpful in penalizing higher-order features.\footnote{We were not able to compare against the method recently proposed in \cite{erion2019learning} due to its prohibitively slow training and large memory requirements.} In fact, using RRR severely decreased performance in all considered metrics, implying that penalizing gradients not only does not help but impedes the learning of relevant features.

\paragraph{Visualizing explanations}

To investigate how CDEP altered a DNN's explanations, we visualize GradCAM heatmaps \citep{uozbulak_pytorch_vis_2019,selvaraju2017grad} on the ISIC test dataset with a regularized and unregularized network in \cref{fig:gradcam}. 
 As expected, after penalizing with CDEP, the DNN attributes less importance to the spurious patches, regardless of their position in the image. More examples are shown in the appendix. 
Anecdotally, patches receive less attribution when the patch color was far from a Caucasian human skin tone, perhaps because these patches are easier for the network to identify. 

\subsection{Combating inductive bias on variants of the MNIST dataset}
\label{subsec:mnist}

In this section, we investigate CDEP's ability to alter which features a DNN uses to perform digit classification, using variants of the MNIST dataset \citep{lecun1998mnist} and a standard CNN architecture for this dataset retrieved from PyTorch \footnote{Retrieved from \href{https://github.com/pytorch/examples/blob/master/mnist/main.py}{\textcolor{cblue}{github.com/pytorch/examples/blob/master/mnist}}.}.

\paragraph{ColorMNIST}
\begin{figure*}[htb!]
	\begin{center}
		\includegraphics[width=\linewidth]{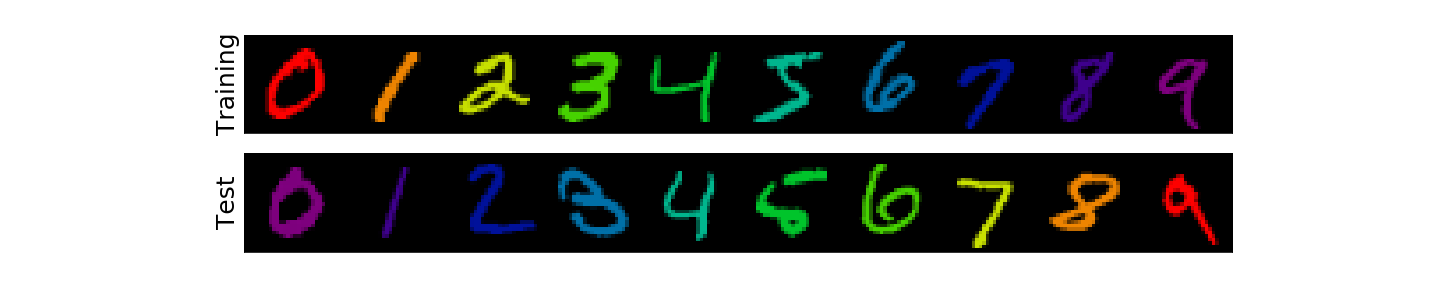}
	\end{center}
	
	\caption{ColorMNIST: the shapes remain the same between the training set and the test set, but the colors are inverted.}
	\label{fig:colorMNIST}
\end{figure*}
Similar to one previous study \citep{li2019repair}, we alter the MNIST dataset to include three color channels and assign each class a distinct color, as shown in \cref{fig:colorMNIST}. An unpenalized DNN trained on this biased data will completely misclassify a test set with inverted colors, dropping to 0\% accuracy (see \cref{tab:mnist}), suggesting that it learns to classify using the colors of the digits rather than their shape.

Here, we want to see if we can alter the DNN to focus on the shape of the digits rather than their color. We stress that this is a toy example where we artificially induced a bias; while the task could be easily solved by preprocessing the input to only have one color channel, this artificial bias allows us to measure the DNN's reliance on the confounding variable \textit{color} in end-to-end training. By design, the task is intuitive and the bias is easily recognized and ignored by humans. However, for a neural network trained in a standard manner, ignoring the confounding variable presents a much greater challenge. 

Interestingly, this task can be approached  by minimizing the contribution of pixels in isolation (which only represent color) while maximizing the importance of groups of pixels (which can represent shapes). To do this, we penalize the CD contribution of sampled single pixel values, following \eref{eq:specific_methods}. By minimizing the contribution of single pixels we encourage the network to focus instead on groups of pixels. Since it would be computationally expensive and not necessary to apply this penalty to every pixel in every training input, we sample pixels to be penalized from the average distribution of nonzero pixels over the whole training set for each batch. 

\cref{tab:mnist} shows that CDEP can partially divert the network's focus on color to also focus on digit shape. We compare CDEP to two previously introduced explanation penalization techniques: penalization of the squared gradients (RRR) \citep{ross2017right} and Expected Gradients (EG) \citep{erion2019learning} on this task. For EG we additionally try penalizing the variance between attributions of the RGB channels (as recommended by the authors of EG in personal correspondence). 
None of the baselines are able to improve the test accuracy of the model on this task above the random baseline, while CDEP is able to significantly improve this accuracy to 31.0\%. We show the increase of predictive accuracy with increasing penalization in the appendix.
Increasing the regularizer rate for CDEP increases accuracy on the test set, implying that CDEP meaningfully captured and penalized the bias towards color.
\begin{table*}[hbt]

\caption{Test Accuracy on ColorMNIST and DecoyMNIST. CDEP is the only method that captures and removes color bias. All values averaged over thirty runs. Predicting at random yields a test accuracy of 10\%.} 
\label{tab:mnist}
\begin{center}
\begin{small}
\begin{sc}
 \begin{tabular}{lccccc}
 \toprule
 & Vanilla & CDEP & RRR & Expected Gradients \\
 \midrule
 \vspace{10pt}
 ColorMNIST & 0.2 $\pm$ 0.2 & \textbf{31.0 $\pm$ 2.3} & 0.2 $\pm$ 0.1 & 10.0 $\pm$ 0.1 \\
 DecoyMNIST & 60.1 $\pm$ 5.1 & \textbf{97.2 $\pm$ 0.8} & \textbf{ 99.0 $\pm$ 1.0 }& \textbf{97.8 $\pm$ 0.2} \\
 \bottomrule
 \end{tabular}
 \end{sc}
\end{small}
\end{center}

\end{table*}

\paragraph{DecoyMNIST}
For further comparison with previous work, we evaluate CDEP on an existing task: DecoyMNIST \citep{erion2019learning}. DecoyMNIST adds a class-indicative gray patch to a random corner of the image. This task is relatively simple, as the spurious features are not entangled with any other feature and are always at the same location (the corners). \cref{tab:mnist} shows that all methods perform roughly equally, recovering the base accuracy. Results are reported using the best penalization parameter $\lambda$, chosen via cross-validation on the validation set. We provide details on the computation time, and memory usage in Table S1, 
showing that CDEP is similar to existing approaches. However, when freezing early layers of a network and finetuning, CDEP very quickly becomes more efficient than other methods in both memory usage and training time.
\subsection{Fixing bias in COMPAS}
\label{ssec:compas}
In all examples so far, the focus has been on improving generalization accuracy. Here,  we turn to improving notions of fairness in models while preserving prediction accuracy instead.

We train and analyze DNNs on the COMPAS dataset \cite{larson2016we}, which contains data for predicting recidivism (i.e whether a person commits a crime / a violent crime within 2 years) from many attributes. Such models have been used for the purpose of informing whether defendants should be incarcerated and can have very serious implication. As a result, we examine and influence the model's treatment of race, restricting our analysis to the subset of people in the dataset whose race is identified as \textit{black} or \textit{white} (86\% of the full dataset). All models were fully connected DNNs with two hidden layers of size 5, ReLU nonlinearity, and dropout rate of 0.1 (see appendix for details).

We analyze the effect of CDEP to alter models with respect to one particular notion of fairness: the wrongful conviction rate (defined as the fraction of defendants who are recommended for incarceration, but did not recommit a crime in the next two years). We aim to keep this rate low and relatively even across races, similar to the common ``equalized odds'' notion of fairness \cite{dieterich2016compas}; note that a full investigation of fairness and its most appropriate definition is beyond the scope of the work here.

\cref{tab:compas} shows results for different models trained on the COMPAS dataset. The first row shows a model trained with standard procedures and the second row shows a model trained with the race of the defendants hidden. 
The unregularized model in the first row has a stark difference in the rates of false positives between \textit{black} and \textit{white} defendants. Black defendants are more than twice as likely to be misclassified as high-risk for future crime. This is in-line with previous analysis of the COMPAS dataset \cite{larson2016we}. 

Obscuring the sensitive attribute from the model does not remove this discrepancy. This is due to the fact that black and white people come from different distributions (e.g. black defendants have a different age distribution). 

The third row shows the results for CDEP, where the model is regularized to place more importance on the race feature and its interactions, encouraging it to learn the dependence between  race and the distribution of other features. By doing so, the model achieves a lower wrongful conviction rate for both black and white defendants, as well as bringing these rates noticeably closer together by disproportionally lowering the wrongful conviction rate for black defendants. Notably, the test accuracy of the model stays relatively fixed despite the drop in wrongful conviction rates.

\begin{table}[H] 
\caption{Fairness measures on the COMPAS dataset. WCR stands for wrongful conviction rate the fraction of innocent defendants who are recommended for incarceration). All values averaged over five runs.} 
\label{tab:compas}
\begin{center}
\begin{small}
\begin{sc}
 \begin{tabularx}{\columnwidth}{llll} 
\toprule
 &   Test acc &  WCR(Black) &  WCR(White)  \\
\midrule
Vanilla         &      67.8$\pm$1.0 &      0.47$\pm$ 0.03 &  0.22$\pm$0.03\\
Race hidden     &     68.5$\pm$0.3   &      0.44$\pm$0.02 &     0.23$\pm$0.01\\
CDEP            &      \textbf{68.8$\pm$0.3}&      \textbf{0.39$\pm$0.04} &  \textbf{0.20$\pm$ 0.01} \\
\bottomrule
\end{tabularx}

 \end{sc}
\end{small}
\end{center}

\end{table}

\subsection{Fixing bias in text data}
\label{subsec:sst}
To demonstrate CDEP's effectiveness on text, we use the Stanford Sentiment Treebank (SST) dataset \citep{socher2013recursive}, an NLP benchmark dataset consisting of movie reviews with a binary sentiment (positive/negative). We inject spurious signals into the training set and train a standard LSTM 
\footnote{Retrieved from \href{https://github.com/clairett/pytorch-sentiment-classificatio}{\textcolor{cblue}{github.com/clairett/pytorch-sentiment-classification}}.} to classify sentiment from the review.

 \begin{figure}[htb]
 \input{sst_example3.tex}
	\caption{Example sentences from the SST dataset with artificially induced bias on gender.}

\label{fig:sst_bias}
\end{figure}

We create three variants of the SST dataset, each with different spurious signals which we aim to ignore (examples in the appendix). In the first variant, we add indicator words for each class (positive: `text', negative: `video') at a random location in each sentence. An unpenalized DNN will focus only on those words, dropping to nearly random performance on the unbiased test set.
In the second variant, we use two semantically similar words (`the', `a') to indicate the class by using one word only in the positive and one only in the negative class. 
In the third case, we use `he' and `she' to indicate class (example in \fref{fig:sst_bias}). Since these gendered words are only present in a small proportion of the training dataset ($\sim2\%$), for this variant, we report accuracy only on the sentences in the test set that do include the pronouns (performance on the test dataset not including the pronouns remains unchanged). \cref{tab:sst_results} shows the test accuracy for all datasets with and without CDEP. In all scenarios, CDEP is successfully able to improve the test accuracy by ignoring the injected spurious signals. 
\begin{table}[!hbtp]
	\caption{Results on SST. CDEP substantially improves predictive accuracy on the unbiased test set after training on biased data.}
	\label{tab:sst_results}
	\begin{center}
	\begin{small}
\begin{sc}
 \begin{tabular}{lrr}
 \toprule
 & Unpenalized & CDEP\\
 \midrule
 Random words & 56.6 $\pm$ 5.8 & \textbf{75.4 $\pm$ 0.9} \\
 Biased (articles) & 57.8 $\pm$ 0.8 & \textbf{68.2 $\pm$ 0.8} \\
 Biased (gender) & 64.2 $\pm$ 3.1 & \textbf{78.0 $\pm$ 3.0} \\
 \bottomrule
 \end{tabular}
 \end{sc}
\end{small}
	\end{center}
\end{table}

\section{Conclusion}
\label{sec:discussion}
In this work we introduce a novel method to penalize neural networks to align with prior knowledge. Compared to previous work, CDEP is the first of its kind that can penalize complex features and feature interactions. Furthermore, CDEP is more computationally efficient than previous work, enabling its use with more complex neural networks.

We show that CDEP can be used to remove bias and improve predictive accuracy on a variety of toy and real data.  The experiments here demonstrate a variety of ways to use CDEP to improve models both on real and toy datasets. CDEP is quite versatile and can be used in many more areas to incorporate the structure of domain knowledge (e.g. biology or physics). The effectiveness of CDEP in these areas will depend upon the quality of the prior knowledge used to determine the explanation targets. 

Future work includes extending CDEP to more complex settings and incorporating more fine-grained explanations and interaction penalizations. We hope the work here will help push the field towards a more rigorous way to use interpretability methods, a point which will become increasingly important as interpretable machine learning develops as a field \citep{doshi2017towards, murdoch2019interpretable}.
\FloatBarrier
\clearpage

\bibliography{refs}
\bibliographystyle{icml2020}
\setcounter{table}{0}
\setcounter{figure}{0}
\setcounter{section}{0}
\renewcommand{\thetable}{S\arabic{table}}
\renewcommand{\thefigure}{S\arabic{figure}}
\renewcommand{\thesection}{S\arabic{section}}

\newpage
\onecolumn

\begin{center}
    \Huge
    Supplement 
\end{center}
\label{sec:supp}

\section{Additional details on MNIST Variants}
\begin{figure}[H]
    \centering
    \includegraphics[width=.2\textwidth]{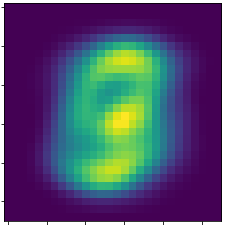}
    \caption{Sampling distribution for ColorMNIST}
\label{fig:colormnist_sample}
\end{figure}

For fixing the bias in the ColorMNIST task, we sample pixels from the distribution of non-zero pixels over the whole training set, as shown in \cref{fig:colormnist_sample}

\begin{figure}[H]
	\begin{center}
		\includegraphics[width=0.4\textwidth]{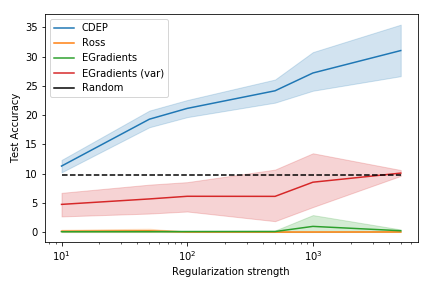}
	\end{center}
	\caption{Results on ColorMNIST (Test Accuracy). All averaged over thirty runs. CDEP is the only method that captures and removes color bias.  }
	\label{fig:colormnist_results}
\end{figure}

For Expected Gradients we show results when sampling pixels as well as when penalizing the variance between attributions for the RGB channels (as recommended by the authors of EG) in \cref{fig:colormnist_results}. Neither of them go above random accuracy, only achieving random accuracy when they are regularized to a constant prediction. 

\subsection{Runtime and memory requirements of different algorithms}
\label{subsec:runtime}

This section provides further details on runtime and memory requirements reported in \tref{tab:decoyMNIST}. We compared the runtime and memory requirements of the available regularization schemes when implemented in Pytorch. 

Memory usage and runtime were tested on the DecoyMNIST task with a batch size of 64. It is expected that the exact ratios will change depending on the complexity of the used network and batch size (since constant memory usage becomes disproportionally smaller with increasing batch size). 

The memory usage was read by recording the memory allocated by PyTorch. Since Expected Gradients and RRR require two forward and backward passes, we only record the maximum memory usage. We ran experiments on a single Titan X.

\begin{table}[H]
	
	\caption{Memory usage and run time for the DecoyMNIST task.}
	\label{tab:decoyMNIST}
	\begin{center}
  \begin{tabular}{lcccc}
  \toprule
   & Unpenalized & CDEP & RRR & Expected Gradients \\
   \midrule

    Run time/epoch (seconds)   & 4.7  & 17.1    & 11.2  & 17.8    \\
    Maximum GPU RAM usage (GB)  & 0.027  & 0.068   & 0.046 & 0.046   \\
    \bottomrule
  \end{tabular}
	\end{center}
\end{table}

\section{Image segmentation for ISIC skin cancer}
\label{subsec:segmentation}
To obtain the binary maps of the patches for the skin cancer task, we first segment the images using SLIC, a common image-segmentation algorithm \citep{achanta2012slic}. Since the patches look quite distinct from the rest of the image, the patches are usually their own segment. 

\begin{figure}[H]
    \centering
    \includegraphics[width=.15\linewidth]{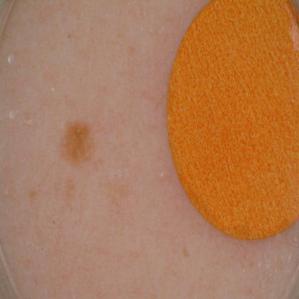}
    \includegraphics[width=.15\linewidth]{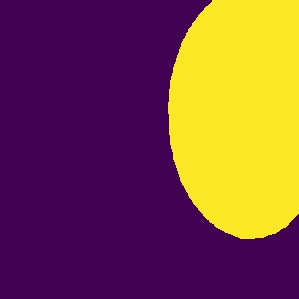}
    \caption{Sample segmentation for the ISIC task.}
\label{fig:image_segmentation}
\end{figure}

Subsequently we take the mean RGB and HSV values for all segments and filtered for segments which the mean was substantially different from the typical caucasian skin tone. Since different images were different from the typical skin color in different attributes, we filtered for those images recursively. As an example, in the image shown in \cref{fig:image_segmentation}, the patch has a much higher saturation than the rest of the image.
For each image we exported a map as seen in \cref{fig:image_segmentation}.

\section{Additional heatmap examples for ISIC}
\label{subsec:additional_heatmaps}
\begin{figure}[H]
    \centering
    \includegraphics[width=.5\textwidth]{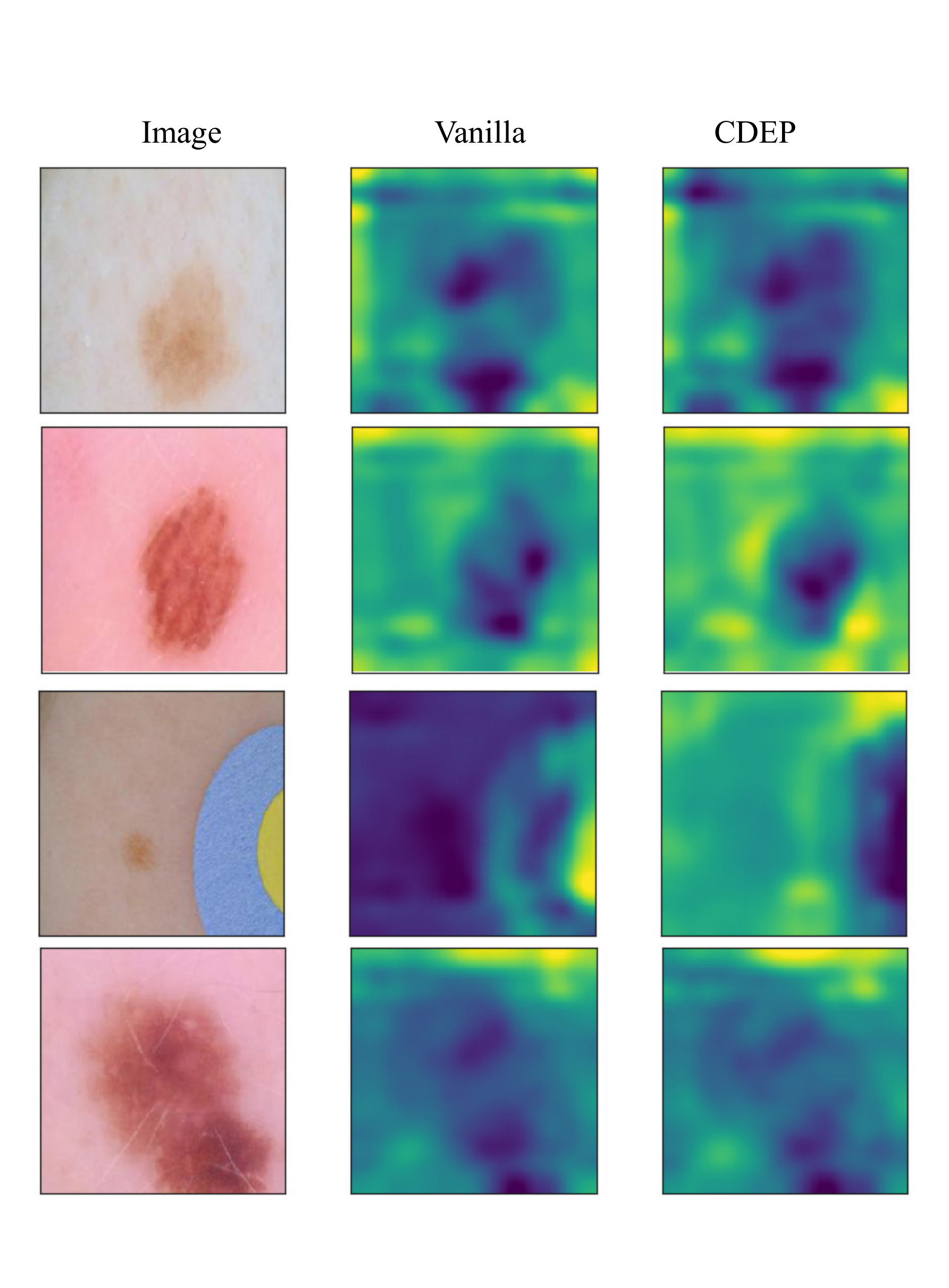}
    \caption{Heatmaps for benign samples from ISIC}
\label{fig:add_examples_benign}
\end{figure}

\begin{figure}[H]
    \centering
    \includegraphics[width=.5\textwidth]{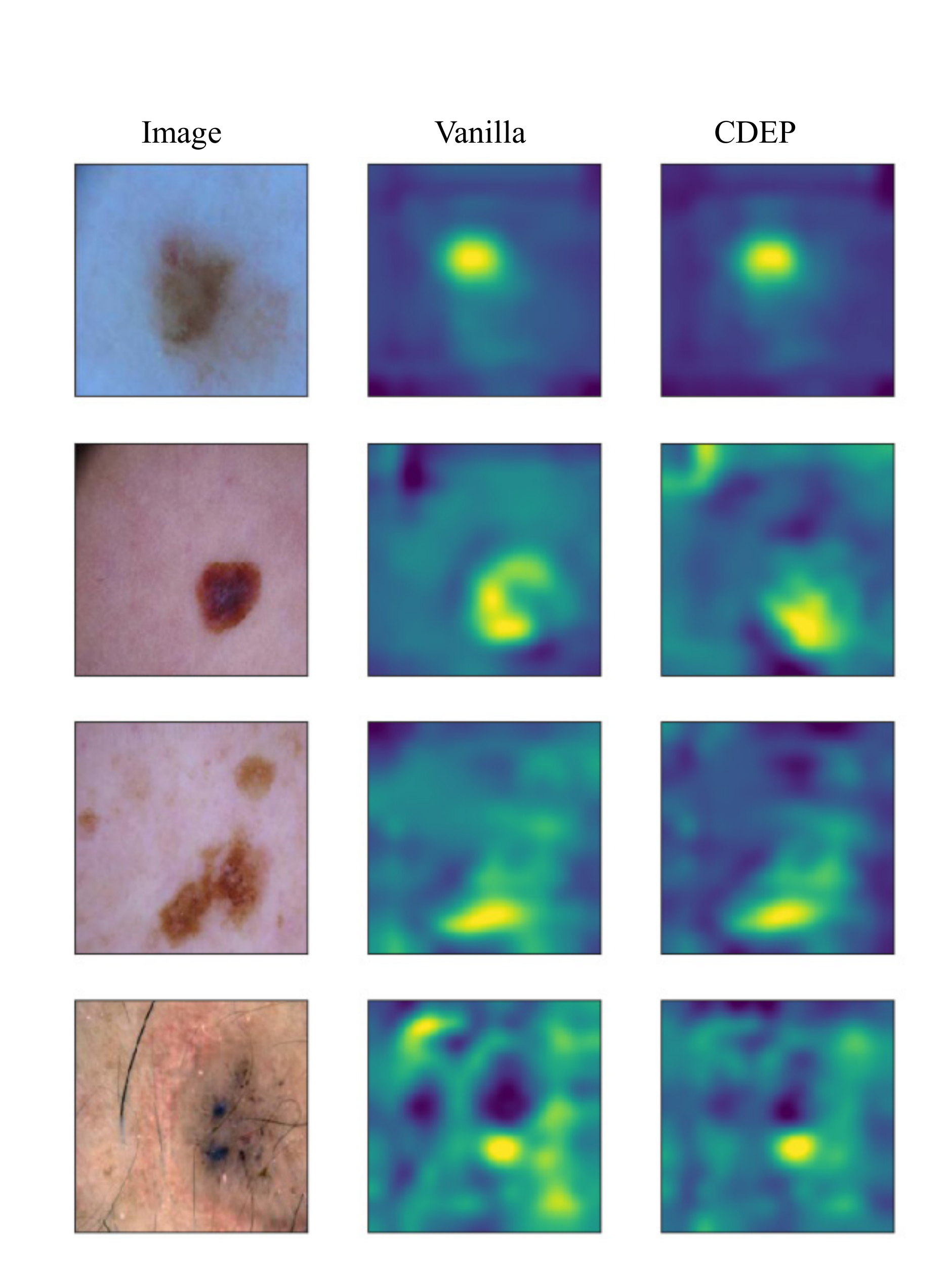}
    \caption{Heatmaps for cancerous samples from ISIC}
\label{fig:add_examples_cancer}
\end{figure}

We show additional examples from the test set of the skin cancer task in \cref{fig:add_examples_benign,fig:add_examples_cancer}. We see that the importance maps for the unregularized and regularized network are very similar for cancerous images and non-cancerous images without patch. The patches are ignored by the network regularized with CDEP. 

A different spurious correlation that we noticed was that proportionally more images showing skin cancer will have a ruler next to the lesion. This is the case because doctors often want to show a reference for size if they diagnosed that the lesion is cancerous. Even though the spurious correlation is less pronounced (in a very rough cursory count, 13\% of the cancerous and 5\% of the benign images contain some sort of measure), the networks learnt to recognize and exploit this spurious correlation. This  further highlights the need for CDEP, especially in medical settings.
\begin{figure}[H]
    \centering
    \includegraphics[width=1.0\textwidth]{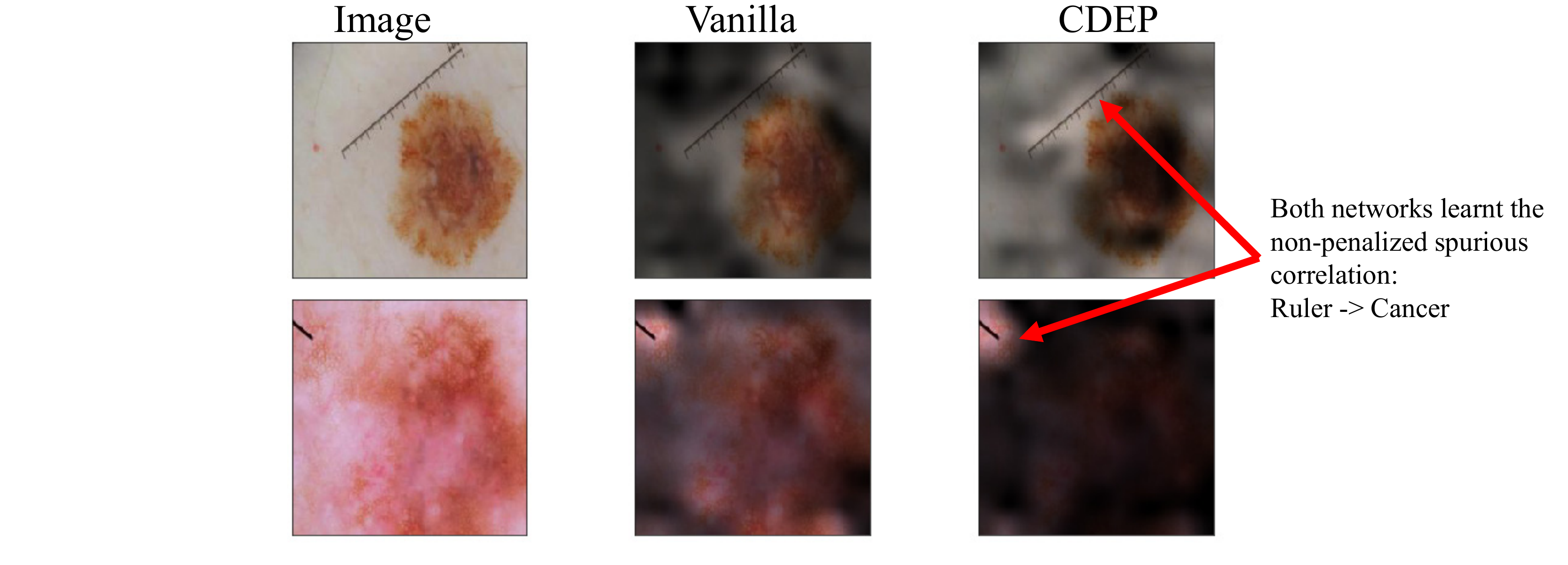}
    \caption{Both networks learnt that proportionally more images with malignant lesions feature a ruler next to the lesion. To make comparison easier, we visualize the heatmap by multiplying it with the image. Visible regions are important for classification.}
\label{fig:ruler_examples}
\end{figure}

\section{Additional details about the COMPAS task}
\label{supp:compas}
In \cref{ssec:compas} we show results for the COMPAS task. For the task, we train a neural network with two hidden layers with five neurons each. The network was trained with SGD (lr 0.01, momentum 0.9), using 0.1 Dropout until loss no longer improved for ten epochs. 

From 7214 samples in the full dataset we excluded 1042 due to missing information about the recidivism as was done in the report from ProPublica \cite{larson2016we}. 

The dataset was preprocessed as follows, following \cite{madras2018predict}. 
\textit{Age} was categorized into <25, 25-45, 45<. \textit{Sex} was categorized into Male/Female. For \textit{Race} we used the given categories, Black, Caucasian, Native American, Other. We also included information about the severity of the crime (Felony/Misdeameanor) as well as the count of previous felonies. Additionally the data included a description of the crime. We parsed this description with matching words and categorized them into  Drugs, Driving, Violence, Robbery and Others as done in \cite{madras2018predict}. As an example, we categorized case descriptions containing 'Battery','Assault', 'Violence' or 'Abuse' into the category Violence.

Since Black and Caucasian are the predominant ethnicities in the dataset, we focus on achieving parity between those two. We excluded other information such as the youth felony count as the proportion of positive samples was very small. 

We split the data into 80\% training , 10\% validation and 10\% test data. 
\section{Additional details about SST task}
\label{subsec:supp_sst}
\cref{subsec:sst} shows the results for CDEP on biased variants of the SST dataset. Here we show examples of the biased sentences (for task 2 and 3 we only show sentences where the bias was present) in \cref{fig:sst1,fig:sst2,fig:sst3}.
For the first task, we insert two randomly chosen words in 100\% of the sentences in the positive and negative class respectively. We choose two words (``text'' for the positive class and ``video'' for the negative class) that were not otherwise present in the data set but had a representation in Word2Vec. 

\begin{figure}[H]

 \input{sst_example1.tex}
    \caption{Example sentences from the variant 1 of the biased SST dataset with decoy variables in each sentence.}
\label{fig:sst1}
\end{figure}

For the second task, we choose to replace two common words ("the" and "a") in sentences where they appear (27\% of the dataset). We replace the words such that one word only appears in the positive class and the other world only in the negative class. By choosing words that are semantically almost replaceable, we ensured that the normal sentence structure would not be broken such as with the first task. 

\begin{figure}[H]

    \input{sst_example2.tex}
    \caption{Example sentences from the variant 2 of the SST dataset with artificially induced bias on articles ("the", "a"). Bias was only induced on the sentences where those articles were used (27\% of the dataset).}
\label{fig:sst2}
\end{figure}

For the third task we repeat the same procedure with two words (``he'' and ``she'') that appeared in only 2\% of the dataset. This helps evaluate whether CDEP works even if the spurious signal appears only in a small section of the data set.

\begin{figure}[H]

   \input{sst_example3.tex}
    \caption{Example sentences from the variant 3 of the SST dataset with artificially induced bias on articles ("he", "she"). Bias was only induced on the sentences where those articles were used (2\% of the dataset).}
\label{fig:sst3}
\end{figure}

\section{Network architectures and training}

For the ISIC skin cancer task we used a pretrained VGG16 network retrieved from the PyTorch model zoo. We use SGD as the optimizer with a learning rate of 0.01 and momentum of 0.9. Preliminary experiments with Adam as the optimizer yielded poorer predictive performance.

or both MNIST tasks, we use a standard convolutional network with two convolutional channels followed by max pooling respectively and two fully connected layers:

Conv(20,5,5) - MaxPool() - Conv(50,5,5) - MaxPool - FC(256) - FC(10). The models were trained with Adam, using a weight decay of 0.001.

Penalizing explanations adds an additional hyperparameter, $\lambda$ to the training. $\lambda $ can either be set in proportion to the normal training loss or at a fixed rate. In this paper we did the latter. We expect that exploring the former could lead to a more stable training process. For all tasks $\lambda$ was tested across a wide range between $[10^{-1}, 10^4]$. 

The LSTM for the SST experiments consisted of two LSTM layers with 128 hidden units followed by a fully connected layer.

\end{document}